\documentclass[11pt]{article}
\usepackage{float}
\usepackage{booktabs}
\usepackage{amsmath}

\usepackage[final]{acl}

\usepackage{times}
\usepackage{latexsym}
\usepackage{forest}
\usepackage{expex}
\usepackage{pgfplots}
\pgfplotsset{compat=1.18}

\usepackage{fvextra}
\DefineVerbatimEnvironment{prompt}{Verbatim}{
  breaklines=true,
  breakanywhere=true,
  fontsize=\small
}

\newenvironment{PromptBox}[1][]{%
  \par\medskip\noindent\begin{minipage}{\linewidth}\hrule\smallskip%
  \if\relax\detokenize{#1}\relax\else #1\par\smallskip\fi%
}{%
  \smallskip\hrule\end{minipage}\par\medskip%
}
\newenvironment{Prompt}{%
  \small\ttfamily\setlength{\parindent}{0pt}%
}{%
}

\usepackage[table]{xcolor}
\definecolor{softpink}{RGB}{255, 230, 240}
\definecolor{strongpink}{RGB}{255, 200, 220}

\lingset{
  aboveexskip=0.6ex,
  belowexskip=0.6ex,
  interpartskip=0.2ex,
  glstyle=nlevel,
  glhangstyle=none,
  everygla=\itshape,  
  everyglb=\upshape,  
  everyglft=\itshape   
}

\usepackage[T1]{fontenc}

\usepackage[utf8]{inputenc}
\DeclareUnicodeCharacter{2248}{\ensuremath{\approx}}

\usepackage{microtype}

\usepackage{inconsolata}

\usepackage{graphicx}

\title{Supervision versus Demonstration-Based In-Context Learning for Multiword Expression Classification}

\author{
  Sercan Karaka\c{s} \\
  University of Chicago \\
  \texttt{skarakas@uchicago.edu}
  \And
  Yusuf \c{S}im\c{s}ek \\
  F{\i}rat University \\
  \texttt{ysimsek@firat.edu.tr}
}

\newcommand{\reponote}{%
\href{https://github.com/ysufsimsek/Supervision-versus-Demonstration-Based-In-Context-Learning-for-Multiword-Expression-Classification}{Code and data repository}%
}

\begin{document}
\maketitle
\begin{abstract}
Turkish idiomatic light verb constructions (LVCs) are challenging for multiword expression processing because they often share the same surface form as fully literal verb–object combinations while functioning as a single, partially idiomatic predicate. We frame Turkish LVC detection as a binary classification task (literal meaning vs. idiomatic meaning) and evaluate on a manually created controlled set (N=147) with matched negatives: out-of-domain random sentences and in-domain literal controls (NLVC), alongside LVC positives. We compare a supervised Turkish encoder baseline (BERTurk with a classifier head) to three instruction-tuned LLMs from different families under zero-shot, one-shot, and few-shot prompting, and analyze how demonstrations shift error profiles. In zero-shot, LLMs perform well on negatives but show very low LVC recall. One-shot prompting sharply improves LVC detection but can induce strong, model-specific biases (over- vs. under-predicting LVC). A richer few-shot prompt improves calibration and yields robust overall performance for GPT-OSS-20B and Qwen 2.5-14B. Overall, the results highlight substantial prompt sensitivity in Turkish metalinguistic classification: the supervised baseline remains competitive, while prompted LLMs can match or exceed it on LVCs with carefully constructed demonstrations. We release code, prompts, and evaluation materials to support reproducibility.\footnote{\reponote}.

\end{abstract}

\section{Introduction}

Multiword expressions (MWEs) are, by definition, sequences of two or more words that behave as a single linguistic unit, often displaying semantic idiosyncrasy, non-compositional meaning, and/or distinctive syntactic behavior \citep{sag2002pain,odijk2013mwe,mititelu-etal-2025-survey-mwe}. Because MWEs are pervasive in everyday language use, they pose persistent challenges for Natural Language Processing (NLP) systems and constitute a well-known hurdle for non-native speakers learning a language \citep{sag2002pain,Ramisch2015, barman-etal-2024-uniformity}. 

Psycholinguistic work indicates that children and adults in both L1 and L2 track distributional statistics of multiword sequences, and that frequent multiword units yield measurable processing advantages \citep{ArnonClark2011,SiyanovaChanturiaConklinVanHeuven2011, CastroviejoEtAl2023}.
If frequent MWEs are entrenched in comprehension and production, as acquisition research suggests \citep{ArnonClark2011,he_wittenberg2020_eventnominals_lvc,schulz2024_mwc_input_variability,chiang2025_light_verbs_mandarin_bilinguals}, then NLP systems intended to model or support language use should handle them reliably across contexts. More broadly, NLP evaluation faces the challenge that many surface forms can realize the same content; automatic metrics are therefore difficult to interpret, and evaluation goals are often under-specified, making human-centered criteria essential \citep{zhou-etal-2022-deconstructing}.
In our setting, this motivates treating human-linguistic distinctions (e.g., whether a verb--nominal sequence forms an idiomatic/light-verb predicate unit vs.\ a literal verb--argument configuration) as a core evaluation target: for downstream Turkish NLP, models should reproduce these basic lexical-semantic generalizations stably across prompting regimes.

To address this, we frame Turkish LVC detection as a binary classification task and evaluate a supervised Turkish encoder baseline (BERTurk) alongside three instruction-tuned LLMs across varying prompting regimes on a manually created controlled set (N=147). Overall results highlight substantial prompt sensitivity : zero-shot LLMs show low LVC recall, whereas carefully constructed few-shot demonstrations improve calibration, allowing prompted LLMs to match or exceed the highly competitive supervised baseline. The remainder of the paper is organized as follows. Section~2 reviews prior work on MWE extraction in general and on Turkish verbal MWEs in particular. Section~3 introduces the present study and motivates the literal--idiomatic contrast used to diagnose LVC detection. Section~4 describes the dataset construction, annotation procedure, and evaluation design. Section~5 presents the supervised BERTurk baselines and the instruction-tuned LLMs evaluated in the study. Section~6 reports the three experiments, comparing zero-shot, one-shot, and few-shot prompting regimes. Section~7 discusses the broader implications of the results, especially the role of prompt sensitivity and task-specific supervision.

\section{Related Work}

\subsection{MWE extraction in general}
A standard distinction in MWE processing is between discovery, which mines candidate expressions from raw text, and identification, which detects and types MWEs in context, typically via supervised tagging or structured prediction. Because many MWEs display substantial syntactic and morphological variability, they cannot be modeled simply as “words with spaces”; effective systems therefore combine lexical and morphosyntactic cues, sometimes in interaction with downstream tasks such as parsing or machine translation \citep{constant-etal-2017-survey}. Alongside work on specific MWE classes such as verb–particle constructions \citep{BaldwinVillavicencio2002}, shared guidelines and multilingual benchmarks have advanced the field, especially through the PARSEME shared task on verbal MWEs, which explicitly targets variability and discontinuity across languages \citep{savary-etal-2017-parseme}. More recently, MWE identification has largely shifted to neural contextual tagging models, from BiLSTM-CRF systems to transformer encoders, often with subword modeling to reduce sparsity and improve generalization across inflectional variants \citep{berk-etal-2018-deep, premasiri-ranasinghe-2022-berts, miletic-walde-2024-semantics}. However, strong token-level performance does not necessarily imply robust semantic modeling. Recent syntheses suggest that transformer-based systems remain inconsistent across settings and may over-rely on surface patterns or memorized lexical cues, especially for idiomatic or semi-lexicalized expressions. Meanwhile, newer PARSEME releases have expanded coverage and improved annotation consistency, enabling more controlled evaluation of discontinuous and morphologically complex VMWEs across languages \citep{savary-etal-2023-parseme}.

\subsection{Turkish}
For Turkish, rich inflection and derivation greatly increase surface variation, making accurate lemmatization and morphosyntactic annotation critical for MWE typing and generalization \citep{oflazer1993twolevel,oflazer-etal-2004-integrating, karakas-simsek-2026-lemmas}; accordingly, early work tightly coupled morphological analysis with MWE processing \citep{oflazer-etal-2004-integrating}. Turkish also has a highly productive inventory of verb--nominal predicates, especially light verb constructions (LVCs), where the nominal contributes core event semantics while the verb is semantically bleached and mainly provides functional material (e.g., event licensing, inflection) \citep{Butt2010LightVerbJungle,Ucar2010LightVerbConstructions,ozbek2010cek}.  Because many light verbs also appear in fully literal transitive uses, the boundary between lexicalized VMWEs and ordinary verb–object combinations is often blurred. Recent work makes use of syntactically annotated resources, including dependency-parsing studies \citep{eryigit-etal-2011-multiword} and treebank-level MWE annotation supporting corpus analysis and supervised identification \citep{eryigit-etal-2015-annotation}. Within PARSEME, Turkish verbal MWEs are explicitly typed (including LVC subclasses), enabling shared evaluation and cross-lingual comparability \citep{savary-etal-2017-parseme}. Still, representation is difficult: UD offers \texttt{compound:lvc}, but deciding whether a nominal is part of a lexicalized predicate (vs. an ordinary object) often requires semantic judgments beyond syntax-only cues \citep{ud-compound-lvc}. Consistently, Turkish corpus work shows that correcting lemma/POS information yields measurable gains in VMWE identification, critically showing how strongly LVC detection depends on robust morphosyntactic preprocessing \citep{ozturk-etal-2022-enhancing} and possibly linguistically-motivated tokenization for large language models \citep{bayram2025tokenizationstandards, bayram2025tokensmeaning}.

Light verb predicates are also theoretically and experimentally informative because they can
decouple surface argument structure from event interpretation. Superficially, many LVCs mirror
ditransitives that encode transfer (e.g., \textit{Ece Murat'a keman-\i\ ver-di} `Ece gave Murat the
violin'), where syntactic arguments map transparently onto giver, theme, and recipient roles.
In LVCs such as \textit{Ece Murat'a \"op\"uc\"uk ver-di} (`Ece gave Murat a kiss'), however, the
same frame admits competing analyses: comprehenders may treat the construction as a
two-participant predicate akin to \textit{Ece Murat'\i\ \"op-t\"u} (`Ece kissed Murat'), or as a
three-role structure in which the nominal behaves like an additional theme
\citep{ozge-etal-2022-assigning}. This mismatch between surface form and predicate meaning is
precisely what makes Turkish LVCs a targeted test case for MWE-aware models.

\section{Present Study}
The central aim of this paper is to investigate Turkish verbal multiword expressions with an explicit focus on the
\textit{literal vs.\ idiomatic} contrast in verb--nominal predicates. MWEs are a well-known
challenge for NLP because they frequently violate surface-based expectations about compositionality,
lexical selection, and distribution, and their behavior varies from semi-compositional to strongly
idiomatic \citep{sag2002pain, baldwinkim2010mwe,constant-etal-2017-survey, savary-etal-2017-parseme}. As we discussed in the previous section, for Turkish
specifically, rich morphology and productive verb--nominal predicate formation make MWE identification
particularly salient \citep{oflazer1993twolevel, oflazer-etal-2004-integrating}.

In addition to standard verbal MWEs, we explicitly target \textit{light verb constructions} (LVCs)
(and closely related idiomatic verb--nominal predicates) in order to reach human standards since human-centered criteria are especially important when defining what counts as success for NLP models \citep{zhou-etal-2022-deconstructing}. 

In these configurations in general and in Turkish, a nominal element contributes the core predicational content while the verb is partially semantically bleached, yielding a single predicate-like unit \citep{grimshaw1988lightverbs, sag2015complex}. Crucially, many Turkish light verbs also occur in fully literal transitive uses, creating hard minimal
contrasts for both humans and NLP models. 

\subsection{Literal--Idiomatic (LVC) Contrast via Lexical Controls}
Our design isolates idiomatic/LVC predication from ordinary verb--argument composition by constructing
matched items where the \emph{same verb} appears in (i) an idiomatic/LVC predicate and (ii) a literal use.
This prevents trivial verb-only heuristics and forces models to attend to whether the verb--nominal sequence
functions as a unitary predicate.

\noindent(1a)\hspace{0.5em}
\begin{tabular}[t]{@{}l@{}}
Ali Ay\c{s}e-ye ilham ver-di.\\
Ali Ay\c{s}e-\textsc{dat} inspiration give-\textsc{pst}-3\textsc{sg}\\
\quad `Ali inspired Ay\c{s}e.' \hfill (\textsc{LVC}; [1])
\end{tabular}

\vspace{0.35em}

\noindent(1b)\hspace{0.5em}
\begin{tabular}[t]{@{}l@{}}
Ali Ay\c{s}e-ye kalem ver-di.\\
Ali Ay\c{s}e-\textsc{dat} pen give-\textsc{pst}-3\textsc{sg}\\
\quad `Ali gave Ay\c{s}e a pen.' \hfill (literal; [0])
\end{tabular}

\noindent Thus, (1a) realizes an LVC in which the nominal \emph{ilham} supplies the core predicate meaning and \emph{ver-} is light, whereas (1b) is a literal transfer event where \emph{ver-} retains its basic ‘give’ meaning and \emph{kalem} is an ordinary theme.

\section{Methods}
For this study, we evaluate (i) a Turkish encoder baseline with a task-specific classifier head (BERTurk) trained on Turkish treebanks data, and (ii) three modern large language models from different model families under three prompting regimes: zero-shot (instruction only), one-shot (instruction + one positive (LVC) and one negative (NLVC) example per target verb template), and few-shot (instruction + a compact set of labeled examples per template). These settings operationalize in-context learning and prompting-based adaptation \citep{brown2020gpt3, liu2023promptsurvey}. 

To train the BERTurk classifier heads, we compiled supervision from Turkish
Universal Dependencies (UD) treebanks distributed in CoNLL-U format:
UD Turkish-Atis, UD Turkish-BOUN, UD Turkish-FrameNet, UD Turkish-GB,
UD Turkish-IMST, UD Turkish-Kenet, UD Turkish-PUD, UD Turkish-Penn,
and UD Turkish-Tourism \citep{ud-tr-atis,ud-tr-boun,ud-tr-framenet,ud-tr-gb,ud-tr-imst,ud-tr-kenet,ud-tr-pud,ud-tr-penn,ud-tr-tourism}.
Each sentence includes POS tags, morphological features, and dependency relations.

To identify candidate LVC realizations from UD annotations, we exploited the
dependency relations \texttt{compound:lvc} and \texttt{compound}
\citep{ud-compound-lvc,ud-compound}.
For treebanks that explicitly annotate LVCs with \texttt{compound:lvc},
we used those arcs directly as LVC candidates.
For treebanks that do not contain \texttt{compound:lvc}, we followed an alternative
procedure: (i) extract all \texttt{compound} dependencies, (ii) retain only those
that form noun--verb dependencies (a nominal dependent linked to a verbal head),
and (iii) manually review these candidates to remove non-LVC cases and to finalize
a linguistically coherent set of target LVC patterns.
Sentences containing validated LVC patterns were labeled as positive ([1]),
and sentences without such patterns were labeled as negative ([0]).

Across the pooled UD data, we started with 82{,}884 sentences. From these, we automatically extracted 10{,}056 candidate LVC sentences using the procedures described above. Importantly, manual verification was applied only to these automatically harvested candidates, not to the full dataset. Two annotators independently reviewed the candidates, and 565 items were identified as mislabeled (266 from the \texttt{compound:lvc}-based extraction and 299 from the \texttt{compound}-based extraction) and removed, leaving 82{,}319 sentences. After filtering, we obtained 9{,}491 LVC instances in total: 1{,}544 derived from \texttt{compound:lvc} arcs and 7{,}947 discovered via the noun--verb \texttt{compound} heuristic.

Across all prompting regimes, we cast the task as binary in-context classification: does a sentence contain an idiomatic/light verb construction (LVC) (\texttt{[1]}) or a literal verb--argument configuration (\texttt{[0]})? To evaluate all models under controlled lexical conditions, we built a bespoke diagnostic dataset of 147 manually authored sentences, split into three balanced conditions (49 each): (i) \textsc{LVC}, (ii) \textsc{NLVC} (literal controls sharing the same target verbs as the LVC items), and (iii) \textsc{Random} (unrelated negative controls). Items were validated by two annotators for naturalness, plausibility, and label correctness. Disagreements were resolved through discussion, and only agreed-upon items were retained in the final dataset. Evaluation items are held out from the one-/few-shot in-context exemplars, and the dataset was created specifically for this study. Annotators guarantee label correctness for the evaluation set, but the BERTurk baseline is trained on UD-derived proxy labels, where negatives may include unannotated LVCs because UD is not necessarily exhaustive.

\begin{figure}[tb]
\centering
\setlength{\abovecaptionskip}{2pt}
\setlength{\belowcaptionskip}{0pt}

\includegraphics[
  height=0.25\textheight,
  keepaspectratio
]{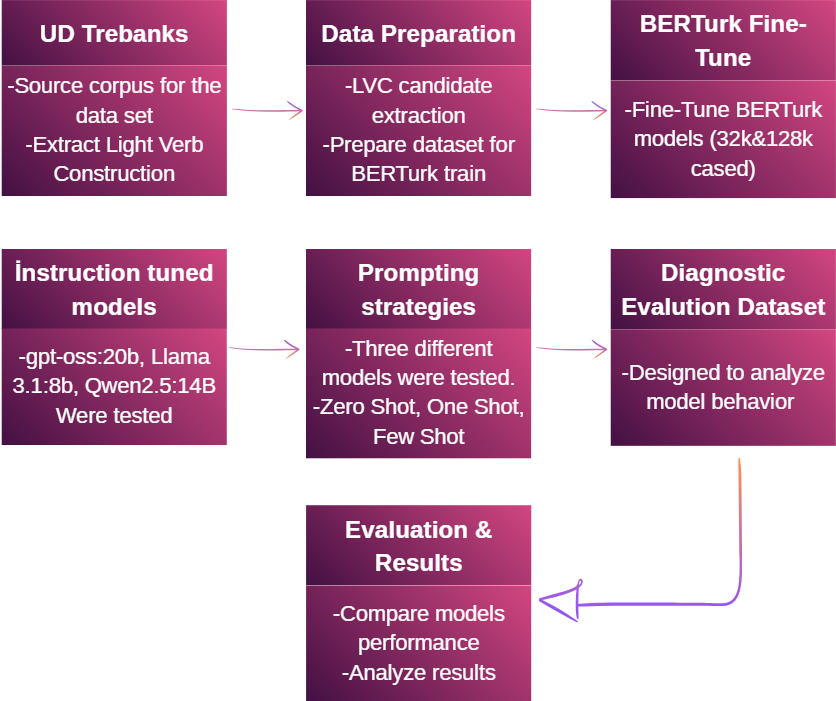}

\caption{Flowchart of the experimental process}
\label{fig:pipeline}
\end{figure}

Although $N=147$ is modest, the dataset is intentionally constructed as a controlled diagnostic set with matched lexical controls. Prior work on behavioral test suites, contrast sets, and dynamic/challenge-style evaluations suggests that small, carefully-designed testbeds can be highly informative for revealing systematic decision-boundary failures that standard i.i.d.\ test accuracy can mask \citep{ribeiro-etal-2020-beyond,gardner-etal-2020-evaluating,kiela-etal-2021-dynabench,yang-etal-2022-testaug,zhao-etal-2024-syntheval,he-etal-2025-minimal-pair-probing,mayne-etal-2025-decision-boundaries,karakas2026benchmarkingsourcesensitivereasoningturkish}. Accordingly, we interpret results primarily as evidence about regime-dependent behavior and calibration under controlled contrasts, rather than as an estimate of broad in-the-wild Turkish LVC detection accuracy.
Figure~\ref{fig:pipeline} presents an overview of the experimental pipeline used in this study.

\section{Models}

We fine-tune BERTurk 32K cased and BERTurk 128K cased \citep{schweter2020berturk} by adding a task-specific binary classification head over the final-layer \texttt{[CLS]} representation. We split the data 80/20 into train/test with stratified sampling and set hidden and attention dropout to 0.2 to reduce overfitting. Models are trained with learning rate $2\times10^{-5}$, batch size 32, weight decay 0.01, for up to 10 epochs, using early stopping on validation loss (patience 3) and selecting the checkpoint with the lowest validation loss. We report Accuracy, F1, and Loss. BERTurk 32K achieves 93.94\% Accuracy, 0.7558 F1, and 0.1500 Loss; BERTurk 128K achieves 94.06\% Accuracy, 0.7508 F1, and 0.1506 Loss, indicating comparable performance across vocabulary sizes.

All LLM experiments were run locally via Ollama. We evaluated instruction-tuned models using the Ollama tags \texttt{llama3.1:8b}, based on the Llama 3.1 model family \citep{grattafiori2024llama3}; \texttt{gpt-oss:20b}, based on OpenAI's gpt-oss model family \citep{openai2025gptoss}; and \texttt{qwen2.5:14b}, based on the Qwen2.5 model family \citep{yang2024qwen25}. The corresponding local Ollama configurations were: \texttt{llama3.1:8b} (architecture \texttt{llama}, 8.0B parameters, quantization \texttt{Q4\_K\_M}, context length 131{,}072), \texttt{gpt-oss:20b} (architecture \texttt{gptoss}, 20.9B parameters, quantization \texttt{MXFP4}, context length 131{,}072), and \texttt{qwen2.5:14b} (architecture \texttt{qwen2}, 14.8B parameters, quantization \texttt{Q4\_K\_M}, context length 32{,}768). The task is formulated as binary classification with labels \texttt{[1]} (contains an LVC) and \texttt{[0]} (no LVC). To improve output consistency, we set \texttt{temperature}=0.1 and left other decoding parameters at their Ollama defaults; each sentence was processed individually, with no batching. Although generation length was not explicitly capped, prompts required outputting only a single label (\texttt{[0]} or \texttt{[1]}). For scoring, outputs were parsed automatically and counted as correct if and only if a valid label was produced; outputs lacking \texttt{[0]}/\texttt{[1]} were treated as invalid and manually inspected ($N=2$).

\section{Experiments}
\subsection{Experiment 1}
Experiment~1 evaluates zero-shot prompting for instruction-tuned LLMs and, for comparison, supervised BERTurk classifier baselines. The task is binary LVC detection under three conditions with $n=49$ items each: Random (negative; unrelated), NLVC (negative; literal in-domain), and LVC (positive; idiomatic/light-verb predicates under our labeling policy). Success rate is computed as correct/$49$ for each condition; Overall pools all $147$ items.

Experiment~1 reveals a strong zero-shot asymmetry: LLMs perform well on negative controls (Random/NLVC) but largely miss LVC positives, consistent with a conservative bias toward [0]. By contrast, supervised BERTurk remains strong on both negatives and LVCs, mitigating the false-negative collapse (Table~\ref{tab:exp1-success}).

\begin{table}[H]
\centering
\small
\setlength{\tabcolsep}{4pt}
\renewcommand{\arraystretch}{1.1}
\rowcolors{2}{softpink}{white}
\begin{tabular}{lcccc}
\hline
\rowcolor{strongpink}
Model & Random & NLVC & LVC & Overall \\
\hline
Llama3.1-8B          & 0.980 & 0.959 & 0.000 & 0.646 \\
GPT-OSS-20B          & 0.939 & 1.000 & 0.061 & 0.667 \\
Qwen2.5-14B          & 0.918 & 0.857 & 0.122 & 0.633 \\
BERTurk-32k (clf)    & 0.980 & 0.816 & 0.673 & 0.823 \\
BERTurk-128k (clf)   & 0.980 & 0.816 & 0.796 & 0.864 \\
\hline
\end{tabular}
\caption{Experiment 1 success rates (0--1). LLM rows are zero-shot prompting; BERTurk rows are supervised classifier baselines (clf). Each condition has $n=49$ items; Overall pools $147$ items.}
\label{tab:exp1-success}
\end{table}

We tested Model$\times$Correctness with chi-square tests per split (reporting Cramer's $V$). Among zero-shot LLMs, differences are not reliable on Random ($\chi^2(2)=2.36$, $p=0.307$, $V=0.13$) but emerge on NLVC ($\chi^2(2)=7.99$, $p=0.018$, $V=0.23$) and LVC ($\chi^2(2)=9.89$, $p=0.007$, $V=0.26$). Including BERTurk yields a very large model effect on LVC ($\chi^2(4)=123.83$, $p=8.13{\times}10^{-26}$, $V=0.71$). Notably, pooling conditions can hide this failure mode: Overall is not significant across LLMs ($\chi^2(2)=0.38$, $p=0.828$) but is across all models ($\chi^2(4)=34.81$, $p=5.07{\times}10^{-7}$).

In Experiment 1, the zero-shot LLMs are extremely conservative: they perform very well on the negative conditions (Random/NLVC) but collapse on the positive LVC condition (especially Llama3.1-8B at 0.000), which suggests a strong bias toward predicting the negative label when no demonstrations are provided. This may additionally reflect limited exposure to Turkish-specific linguistic patterns, as Llama 3.1 is not explicitly trained or optimized for Turkish.
This creates a misleadingly “decent” Overall accuracy despite near-total failure on the phenomenon of interest (LVCs). By contrast, the supervised BERTurk classifier baselines maintain high LVC success (0.673–0.796) while staying strong on negatives, indicating that task-specific Turkish supervision captures the light-verb/idiom signal that zero-shot prompting does not. Statistically, model differences are driven primarily by the LVC condition (large effect when including BERTurk), while Overall differences among the three LLMs alone are not significant, which again shows why condition-wise reporting matters.

Although BERTurk has far fewer parameters than the instruction-tuned LLMs we test, it is evaluated here with a task-specific classifier head trained with labeled supervision derived from Turkish treebank resources. This kind of supervision may make certain morphosyntactic regularities more directly exploitable for a metalinguistic decision boundary (LVC vs.\ literal), even when overall model capacity is smaller. More generally, prior work on pretrained encoders suggests that additional supervised training---either as intermediate-task fine-tuning or task-specific fine-tuning---can sometimes improve downstream generalization and make particular linguistic distinctions more accessible to a simple classifier \citep{phang2018stilts,pruksachatkun2020intermediate}. Related analyses also indicate that contextual encoders can encode multi-level linguistic information that downstream heads can leverage \citep{tenney2019bert}. Since treebank annotation frameworks (e.g., Universal Dependencies) explicitly target predicate--argument structure and may include dedicated analyses for complex predicates (including light-verb-like patterns in some languages) \citep{nivre2016ud,nivre2020ud2}, supervised exposure to such representations might partially account for why a smaller encoder+head baseline can appear comparatively strong on this metalinguistic classification task. However, because the supervision signal is not identical to our labeling policy, and because training data and evaluation items can differ in domain and lexical coverage, we treat this explanation as suggestive rather than causal, which requires further research.
\subsection{Experiment 2: One-per-Class Prompting}

In Experiment~2, we use one labeled example for LVC and NLVC per target verb template  in-context demonstration along with instruction, and then require the model to output only a binary label for each test sentence in our evaluation corpus.

The one-shot prompt yields sharply different error profiles across model families (Table~\ref{tab:exp2-oneshot}). Llama 3.1 8B attains high accuracy on LVC items but performs poorly on both negative splits, consistent with over-predicting the positive label. Qwen 2.5 shows the opposite bias: near-ceiling performance on negatives but substantially lower accuracy on LVC positives. GPT-OSS-20B is comparatively more balanced across conditions.

\begin{table}[H]
\centering
\small
\setlength{\tabcolsep}{4pt}
\rowcolors{2}{softpink}{white}
\begin{tabular}{lcccc}
\hline
\rowcolor{strongpink}
Model & LVC & NLVC & Random & Overall \\
\hline
GPT-OSS-20B   & 0.837 & 0.735 & 0.898 & 0.823 \\
Llama 3.1 8B  & 0.878 & 0.286 & 0.469 & 0.544 \\
Qwen 2.5 14B  & 0.490 & 1.000 & 0.959 & 0.816 \\
\hline
\end{tabular}
\caption{Experiment~2 success rates. Each split has $n=49$ items; Overall pools $N=147$ items.}
\label{tab:exp2-oneshot}
\end{table}

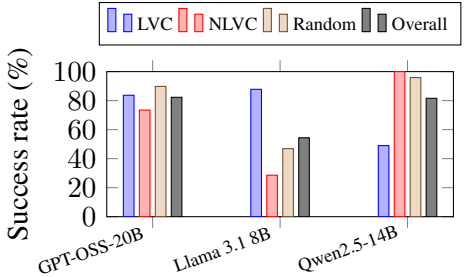
\begin{figure}[H]
\centering
\begin{tikzpicture}
\begin{axis}[
  ybar,
  bar width=4pt,
  width=0.8\columnwidth,
  height=3.5cm,
  ymin=0, ymax=100,
  ylabel={Success rate (\%)},
  symbolic x coords={GPT-OSS-20B,Llama 3.1 8B,Qwen2.5-14B},
  xtick=data,
  x tick label style={rotate=20,anchor=east,font=\scriptsize},
  legend style={font=\scriptsize, at={(0.5,1.18)}, anchor=south, legend columns=4},
  enlarge x limits=0.18
]
\addplot coordinates {(GPT-OSS-20B,83.7) (Llama 3.1 8B,87.8) (Qwen2.5-14B,49.0)};
\addplot coordinates {(GPT-OSS-20B,73.5) (Llama 3.1 8B,28.6) (Qwen2.5-14B,100.0)};
\addplot coordinates {(GPT-OSS-20B,89.8) (Llama 3.1 8B,46.9) (Qwen2.5-14B,95.9)};
\addplot coordinates {(GPT-OSS-20B,82.3) (Llama 3.1 8B,54.4) (Qwen2.5-14B,81.6)};
\legend{LVC,NLVC,Random,Overall}
\end{axis}
\end{tikzpicture}
\caption{Experiment~2 (one-shot) success rates by condition and overall (percent).}
\label{fig:exp2-bar}
\end{figure}

To test whether models differ reliably within each split, we apply chi-square tests of independence (Model $\times$ Correctness) and use a Holm correction across splits. All splits show significant model differences after correction (LVC: $\chi^2(2)=22.825$, $p_{\text{Holm}}=1.106\times 10^{-5}$; NLVC: $\chi^2(2)=58.095$, $p_{\text{Holm}}=9.704\times 10^{-13}$; Random: $\chi^2(2)=40.091$, $p_{\text{Holm}}=5.909\times 10^{-9}$; Overall: $\chi^2(2)=37.574$, $p_{\text{Holm}}=1.386\times 10^{-8}$).

Post-hoc two-proportion tests (Holm-corrected within split) indicate that on LVC items, GPT-OSS-20B and Llama 3.1 8B outperform Qwen 2.5, whereas on negative splits, Qwen 2.5 and GPT-OSS-20B outperform Llama 3.1 8B; overall, GPT-OSS-20B and Qwen 2.5 are statistically indistinguishable and both outperform Llama 3.1 8B.

In general, the one-shot regime changes models’ decision thresholds rather than uniformly improving “LVC understanding,” producing clear bias trade-offs across families. Llama~3.1~8B appears to over-predict the positive label, which boosts LVC hit rate but collapses performance on both negative splits, while Qwen~2.5 shows the opposite pattern—near-ceiling negatives but substantially weaker LVC detection, suggesting a conservative, negative-skewed classifier. GPT-OSS-20B is the most balanced under this prompt, maintaining strong negative performance while still achieving relatively high LVC accuracy. Overall, the results imply that a single demonstration can induce strong, model-specific calibration shifts, so one-shot prompting is not reliably “better” without checking per-condition error profiles.

\subsection{Experiment 3: Few-shot prompting}

Experiment~3 evaluates the same three instruction-tuned LLMs under a few-shot prompt
that provides multiple in-context demonstrations (one positive and one negative for several verbs) and constrains outputs to a binary label (\texttt{[1]} for LVC, \texttt{[0]} otherwise).

Across models, the few-shot prompt yields broadly high overall accuracy for GPT-OSS-20B and
Qwen 2.5 (84--86\%), while Llama 3.1 8B remains lower overall due to substantially weaker performance
on the negative conditions despite strong LVC accuracy. Model differences are robust on Random and NLVC
(and overall) under chi-square tests with Holm correction, but the omnibus cross-model difference on LVC
items is weaker after correction, suggesting more similar positive-class performance under this prompt.
The error profiles remain asymmetric: Qwen 2.5 is relatively conservative (very strong negatives but
more missed LVCs), whereas Llama 3.1 8B is comparatively liberal (high LVC hit-rate but many false positives);
GPT-OSS-20B is the most balanced of the three.

\begin{table}[H]
\centering
\small
\setlength{\tabcolsep}{3pt}
\renewcommand{\arraystretch}{1.1}
\rowcolors{2}{softpink}{white}
\begin{tabular}{lcccc}
\hline
\rowcolor{strongpink}
Model & Random & NLVC & LVC & Overall \\
\hline
GPT-OSS-20B    & 91.8 & 75.5 & 85.7 & 84.4 \\
Llama 3.1 8B   & 51.0 & 61.2 & 87.8 & 66.7 \\
Qwen 2.5 14B   & 87.8 & 98.0 & 71.4 & 85.7 \\
\hline
\end{tabular}
\caption{Experiment~3 success rates (\%). Each condition has $n=49$ items; Overall pools $N=147$.}
\label{tab:exp3-fewshot}
\end{table}


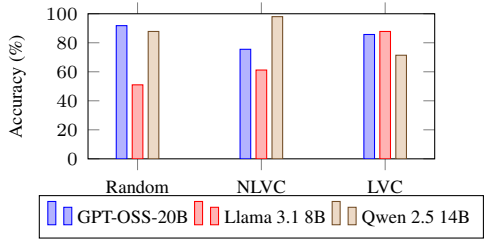
\begin{figure}[H]
\centering
\begin{tikzpicture}
\begin{axis}[
    ybar,
    width=0.8\columnwidth,
    height=3.5cm,
    ymin=0, ymax=100,
    ylabel={Accuracy (\%)},
    symbolic x coords={Random,NLVC,LVC},
    xtick=data,
    bar width=4pt,
    enlarge x limits=0.20,
    legend style={at={(0.5,-0.25)},anchor=north,legend columns=3,font=\scriptsize},
    tick label style={font=\scriptsize},
    label style={font=\scriptsize},
]
\addplot coordinates {(Random,91.8) (NLVC,75.5) (LVC,85.7)};
\addplot coordinates {(Random,51.0) (NLVC,61.2) (LVC,87.8)};
\addplot coordinates {(Random,87.8) (NLVC,98.0) (LVC,71.4)};
\legend{GPT-OSS-20B, Llama 3.1 8B, Qwen 2.5 14B}
\end{axis}
\end{tikzpicture}
\caption{Experiment~3 accuracy by condition (few-shot prompting).}
\label{fig:exp3-bars}
\end{figure}

We tested cross-model differences per split using chi-square independence tests (Model$\times$Correctness) with Holm correction over Random/NLVC/LVC/Overall. Effects are significant for Random ($\chi^2(2)=27.85$, $p_{\text{Holm}}=4.0{\times}10^{-6}$, $V=0.44$), NLVC ($\chi^2(2)=19.73$, $p_{\text{Holm}}=1.4{\times}10^{-4}$, $V=0.37$), and Overall ($\chi^2(2)=19.95$, $p_{\text{Holm}}=1.4{\times}10^{-4}$, $V=0.21$), but not for LVC after correction ($\chi^2(2)=5.17$, $p_{\text{Holm}}=0.075$, $V=0.19$). Post-hoc two-proportion tests (Holm-corrected within split) show that on Random, GPT-OSS-20B and Qwen~2.5 outperform Llama~3.1~8B, while GPT-OSS-20B and Qwen~2.5 do not differ; on NLVC, Qwen~2.5 outperforms both GPT-OSS-20B and Llama~3.1~8B, with no reliable GPT-OSS-20B vs.\ Llama difference; and on LVC, no pairwise differences survive correction. Finally, condition effects within models are significant for Llama~3.1~8B ($\chi^2(2)=15.86$, $p_{\text{Holm}}=0.0011$) and Qwen~2.5 ($\chi^2(2)=14.33$, $p_{\text{Holm}}=0.0015$), but not for GPT-OSS-20B ($\chi^2(2)=5.05$, $p=0.080$), consistent with GPT-OSS-20B's more even profile.

\definecolor{softpink}{RGB}{255, 230, 240}
\definecolor{strongpink}{RGB}{255, 180, 210}

\begin{table*}[t]
  \centering
  \small
  \setlength{\tabcolsep}{3.5pt}
  \renewcommand{\arraystretch}{0.78}
  \rowcolors{2}{softpink}{white}
  \begin{tabular}{llccccrrcc}
    \hline
    \rowcolor{strongpink}
    Model & Regime & Random & NLVC & LVC & Overall & FP & FN & Prec & Rec \\
    \hline
    \multicolumn{10}{l}{GPT-OSS-20B} \\
    \hspace{1em} & Zero-shot & 93.9 & 100.0 & 6.1  & 66.7 &  3 & 46 & 50.0 & 6.1 \\
    \hspace{1em} & One-shot  & 89.8 & 73.5  & 83.7 & 82.3 & 18 &  8 & 69.5 & 83.7 \\
    \hspace{1em} & Few-shot (master) & 91.8 & 75.5 & 85.7 & 84.4 & 16 &  7 & 72.4 & 85.7 \\
    \hline
    \multicolumn{10}{l}{Qwen 2.5 14B} \\
    \hspace{1em} & Zero-shot & 91.8 & 85.7 & 12.2 & 63.3 & 11 & 43 & 35.3 & 12.2 \\
    \hspace{1em} & One-shot  & 95.9 & 100.0 & 49.0 & 81.6 &  2 & 25 & 92.3 & 49.0 \\
    \hspace{1em} & Few-shot (master) & 87.8 & 98.0 & 71.4 & 85.7 &  7 & 14 & 83.3 & 71.4 \\
    \hline
    \multicolumn{10}{l}{Llama 3.1 8B} \\
    \hspace{1em} & Zero-shot & 98.0 & 95.9 & 0.0  & 64.6 &  3 & 49 & 0.0  & 0.0 \\
    \hspace{1em} & One-shot  & 46.9 & 28.6 & 87.8 & 54.4 & 61 &  6 & 41.3 & 87.8 \\
    \hspace{1em} & Few-shot (master) & 51.0 & 61.2 & 87.8 & 66.7 & 43 &  6 & 50.0 & 87.8 \\
    \hline
    \multicolumn{10}{l}{Supervised baselines} \\
    BERTurk-32k (clf)  & Supervised & 98.0 & 81.6 & 67.3 & 82.3 & 10 & 16 & 76.7 & 67.3 \\
    BERTurk-128k (clf) & Supervised & 98.0 & 81.6 & 79.6 & 86.4 & 10 & 10 & 79.6 & 79.6 \\
    \hline
  \end{tabular}
  \caption{\label{tab:prompt-regimes}
    Prompt regime comparison across Experiments~1--3.
    Entries are success rates (\%) for each condition (each has $n=49$) and Overall ($N=147$).
    FP/FN are pooled error counts where FP counts mistakes on negatives (Random+NLVC) and FN counts mistakes on LVC.
    Precision/Recall treat LVC as the positive class and pool Random+NLVC as negatives.
  }
\end{table*}

In general, experiment 3 (few-shot) largely eliminates the zero-shot “always-negative” failure mode, yielding high overall accuracy for GPT-OSS-20B and Qwen 2.5 (≈84–86 per cent), while Llama 3.1 8B remains substantially lower overall because it still over-predicts the positive label on the negative splits. The remaining differences are driven mainly by negative conditions: Qwen 2.5 is near-ceiling on NLVC/Random but misses more LVCs, whereas Llama shows the opposite bias (high LVC hit-rate, many false positives). GPT-OSS-20B is the most balanced profile across splits, and statistically, cross-model differences are robust for Random/NLVC (and overall) but noticeably weaker for LVC, suggesting that under a sufficiently rich prompt, models converge more on the positive class than on calibrating negatives.

\section{General Discussion}
Table~\ref{tab:prompt-regimes} summarizes how prompt regime modulates Turkish LVC detection. Across LLMs, zero-shot performance is dominated by missed positives (high FN), while adding demonstrations mainly shifts the decision boundary and can flip the dominant error type (e.g., Llama becomes overly positive in one-shot). The few-shot prompt generally improves calibration by exposing broader positive/negative variability, reducing extreme one-shot failure modes while preserving high LVC hit-rates. For context, supervised BERTurk (classifier head) is more balanced in Experiment~1, though it is not directly comparable to prompted LLMs due to task-specific training.

Consistent with this, zero-shot cross-model differences are limited: after Holm correction, only NLVC shows a reliable Model$\times$Correctness association ($\chi^2(2)=6.45$, $p_{\text{Holm}}=0.040$, $V=0.26$), while Random and Overall do not ($p_{\text{Holm}}\ge 0.40$) and the LVC omnibus is not reliable ($\chi^2(2)=6.39$, $p_{\text{Holm}}=0.123$). In contrast, one-shot yields robust cross-model differences on every split (all $p_{\text{Holm}}\le 1.11{\times}10^{-5}$). Under the few-shot prompt, cross-model differences remain reliable for Random ($\chi^2(2)=27.85$, $p_{\text{Holm}}=3.58{\times}10^{-6}$, $V=0.44$), NLVC ($\chi^2(2)=19.73$, $p_{\text{Holm}}=1.40{\times}10^{-4}$, $V=0.37$), and Overall ($\chi^2(2)=19.95$, $p_{\text{Holm}}=1.40{\times}10^{-4}$, $V=0.21$), but not for LVC after correction ($\chi^2(2)=5.17$, $p_{\text{Holm}}=0.075$), consistent with partial convergence on positives under the prompt.

Within each model, prompt regime has a strong effect on LVC decisions. Regime$\times$Correctness tests across the three prompting regimes show large and reliable regime effects on LVC for GPT-OSS-20B ($\chi^2(2)=68.96$, $p_{\text{Holm}}=4.25\times 10^{-15}$, $V=0.60$) and Qwen~2.5 ($\chi^2(2)=35.48$, $p_{\text{Holm}}=7.90\times 10^{-8}$, $V=0.42$). For Llama~3.1~8B, LVC accuracy jumps from zero-shot to prompted regimes (near-ceiling recall thereafter), but this co-occurs with a large rise in false positives on negatives (Table~\ref{tab:prompt-regimes}). Taken together, these regime-dependent FP/FN trade-offs are compatible with the view that in-context demonstrations primarily re-weight a model's decision boundary and calibration, that is, they change how aggressively the model predicts the positive class, rather than uniformly improving performance across splits \citep{zhao-etal-2021-calibrate,min-etal-2022-rethinking,liu2023promptsurvey,akyurek2024icl}. In this sense, few-shot prompting appears to elicit metalinguistic judgments about LVC vs.\ literal usage, while the large regime effects also highlight substantial prompt sensitivity and contextual/label bias documented in prior work on in-context learning \citep{zhao-etal-2021-calibrate,min-etal-2022-rethinking}.

At first sight, the stronger performance of the fine-tuned BERT model over several larger recent LLMs may seem surprising. We do not interpret this result as evidence that BERT is generally more capable than state-of-the-art LLMs. However, we think that it reflects the difference between in-domain supervised sequence labeling and general-purpose prompting. Idiomatic light verb identification is a fixed, annotation-scheme-sensitive token classification problem: the model must decide whether a verb participates in a construction such as a noun plus \textit{et-} `do', \textit{ol-} `be/become', or \textit{yap-} `do/make' under the conventions of the annotated dataset. A fine-tuned encoder is directly optimized for this label space and can exploit recurring morphosyntactic, lexical, and positional cues. Prompted LLMs, by contrast, must infer the task definition from an instruction and map open-ended linguistic knowledge onto a discrete annotation decision without task-specific parameter updates. This exposes them to additional sources of error, including output-format instability, boundary mismatches, inconsistent treatment of borderline idiomaticity, and sensitivity to prompt wording. This interpretation is consistent with broader findings that task-specific supervision or parameter-efficient adaptation can match or outperform zero-/few-shot prompting and in-context learning, especially in narrow classification settings \citep{mosbach-etal-2023-shot,liu2022fewshot,edwards-camacho-collados-2024-language,bucher2024finetuned,pecher2024finetuning}. Recent work also shows that prompted and in-context learning approaches remain sensitive to demonstration selection, number, order, and even the position of examples within the prompt \citep{schoch-ji-2025-monte,gao-etal-2025-learning,cobbina-zhou-2025-show}. Thus, the relevant contrast in our experiments is not ``BERT versus LLMs'' in general, but supervised in-domain token-level classification versus prompted general-purpose inference. For a fine-grained morphosyntactic and idiomaticity-sensitive annotation task, direct supervision over the target label space can be more valuable than broad instruction-following ability.

BERTurk uses task-specific supervised training (a classifier head with labeled supervision), whereas the LLMs are evaluated via prompting without gradient updates; as a result, direct comparison may not be reliable. Still, the results suggest two descriptive trends: (i) relative to GPT-OSS-20B in zero-shot, BERTurk-128k achieves much higher LVC accuracy (0.796 vs.\ 0.061; two-proportion $z=7.35$, $p=2.0\times 10^{-13}$) and higher Overall accuracy (0.864 vs.\ 0.667; $z=3.99$, $p=6.6\times 10^{-5}$); (ii) once GPT-OSS-20B is provided demonstrations (one-/few-shot), its Overall performance becomes closer to BERTurk-128k (e.g., 0.844 vs.\ 0.864 in Experiment~3; $p=0.62$), and it may even slightly exceed BERTurk on LVC positives (0.857 vs.\ 0.796; $p=0.42$), though these differences are not statistically reliable at our sample size. One possible interpretation is that explicit supervised exposure to Turkish morphosyntax and annotation conventions may help stabilize metalinguistic judgments, while prompted LLMs can approximate this behavior when the prompt supplies sufficient labeled structure; prior work suggests supervised objectives can sharpen linguistically relevant representations, but the mapping from such supervision to metalinguistic competence is not guaranteed and likely task-dependent \citep{tenney2019bert,hewitt2019structural,rogers2020bertology}.

\section{Conclusion}
This paper examined Turkish LVC detection through a literal–idiomatic contrast that blocks trivial verb-only heuristics and isolates whether models treat verb–nominal predicates as unitary multiword meanings. Across three prompting regimes, we find that instruction-tuned LLMs are conservative in zero-shot (high negative accuracy, near-collapse on LVC recall), but can be rapidly re-calibrated with minimal demonstrations, albeit with distinct family-specific biases in one-shot. A few-shot prompt generally stabilizes behavior and yields strong overall performance, with GPT-OSS-20B remaining the most balanced across conditions and Qwen 2.5 excelling on negatives. At the same time, the supervised BERTurk classifier provides a strong Turkish baseline that is competitive overall, suggesting that task-specific supervision may still offer advantages for metalinguistic decisions in morphologically rich languages. Thus, the findings appear to motivate treating LVC detection as a prompt-sensitive capability and show the value of controlled, condition-wise evaluation beyond aggregate accuracy.

\section*{Limitations}

Our evaluation focuses on a targeted set of Turkish verb–nominal predicates (Random/NLVC/LVC; $N=147$), so the conclusions may not fully generalize to other Turkish MWE families (e.g., fixed idioms, postpositional MWEs) or to broader domains beyond the curated test set. In addition, prompted LLM performance is sensitive to prompt design and demonstration choice, so reported one-/few-shot results should be interpreted as evidence about \emph{prompt-regime effects} rather than model-intrinsic competence. From a mechanistic-interpretability perspective, prompts can be viewed as causal interventions on internal computation, and linking prompt-induced behavior shifts to stable, abstracted mechanisms remains an open challenge \citep{andreas2022agent, BereskaGavves2024,GeigerEtAl2025, holtzman2025prompting}. Finally, BERTurk uses supervised training with a classifier head, whereas LLMs are evaluated via in-context prompting; we therefore treat the cross-family comparison as suggestive rather than strictly like-for-like.

\bibliography{custom}

\clearpage
\appendix

\clearpage
\appendix
\onecolumn

\clearpage
\appendix
\onecolumn

\section{Prompt Templates}
\label{app:prompts}

\subsection{Zero-shot}
\label{app:prompts-zero}

\begin{PromptBox}{\textbf{Zero-shot}}
\begin{Prompt}
Görev: Aşağıdaki cümlede “light verb construction (LVC)” var mı?

Cümle: ``{sentence}''

Tanım:
\begin{itemize}
    \item[\textbf{[1]}] LVC varsa: Yüklemin çekirdek anlamını bir isim taşır; fiil daha “hafif/yardımcı” rol oynar ve yapı tek bir yüklem gibi çalışır.  
    Not: Bu deneyde kalıplaşmış/idiomatik isim+fiil yüklemlerini de [1] say.
    \item[\textbf{[0]}] LVC yoksa: Fiil kendi temel (literal) anlamıyla kullanılıyordur (fiziksel eylem, nesne aktarma, gerçek görme/duyma vb.).
\end{itemize}

Eğer cümlede “light verb construction (LVC)” varsa sadece [1] yaz.  
Eğer cümlede “light verb construction (LVC)” yoksa sadece [0] yaz.  

Başka açıklama yapma, sadece sonucu yaz.

Cevap:
\end{Prompt}
\end{PromptBox}

\subsection{One-shot}
\label{app:prompts-one}

\begin{PromptBox}{\textbf{One-shot}}
\begin{Prompt}
Görev: Aşağıdaki cümlede “light verb construction (LVC)” var mı?

Tanım:
\begin{itemize}
    \item[\textbf{[1]}] LVC varsa: Yüklemin çekirdek anlamını bir isim taşır; fiil daha “hafif/yardımcı” rol oynar ve yapı tek bir yüklem gibi çalışır.  
    Not: Bu deneyde kalıplaşmış/idiomatik isim+fiil yüklemlerini de [1] say.
    \item[\textbf{[0]}] LVC yoksa: Fiil kendi temel (literal) anlamıyla kullanılıyordur (fiziksel eylem, nesne aktarma, gerçek görme/duyma vb.).
\end{itemize}

Örnekler (fiile göre):

\textbf{[VERB-\textit{DUY}]}

Cümle: ``Ona büyük saygı duydu.''  
Cevap: [1]

Cümle: ``Koridordan gelen sesi duydu.''  
Cevap: [0]

\vspace{0.5em}
Şimdi sınıflandır:

Cümle: ``{sentence}''

Yalnızca [0] ya da [1] yaz. Başka hiçbir şey yazma.

Cevap:
\end{Prompt}
\end{PromptBox}

\subsection{Master few-shot (example; abbreviated)}
\label{app:prompts-master-example}

\begin{PromptBox}{\textbf{Master few-shot (example; abbreviated)}}
\begin{Prompt}
Görev: Aşağıdaki cümlede “light verb construction (LVC)” var mı?

Tanım:
\begin{itemize}
    \item[\textbf{[1]}] LVC varsa: Yüklemin çekirdek anlamını bir isim taşır; fiil daha “hafif/yardımcı” rol oynar ve yapı tek bir yüklem gibi çalışır.  
    Not: Bu deneyde kalıplaşmış/idiomatik isim+fiil yüklemlerini de [1] say.
    \item[\textbf{[0]}] LVC yoksa: Fiil kendi temel (literal) anlamıyla kullanılıyordur (fiziksel eylem, nesne aktarma, gerçek görme/duyma vb.).
\end{itemize}

Örnekler (kısaltılmış; temsili örnek):

\textbf{[\textit{DUY-}]}

Cümle: ``Hayatında ilk defa birine güven duydu.''  
Cevap: [1]

Cümle: ``Koridordan gelen sesleri duydu.''  
Cevap: [0]

\medskip
\textbf{[\textit{VER-}]}

Cümle: ``Soruma hemen yanıt verdi.''  
Cevap: [1]

Cümle: ``Öğretmen öğrencilere kâğıt verdi.''  
Cevap: [0]

\medskip
\textbf{[\textit{GİR-}]}

Cümle: ``Sınav başlayınca paniğe girdi.''  
Cevap: [1]

Cümle: ``Odaya hızlıca girdi.''  
Cevap: [0]

\medskip
... (diğer fiiller ve örnek çiftleri aynı formatta devam eder)

\vspace{0.5em}
Şimdi sınıflandır:

Cümle: ``{sentence}''

Yalnızca [0] ya da [1] yaz. Başka hiçbir şey yazma.

Cevap:
\end{Prompt}
\end{PromptBox}

\section{English Translations (instructions only)}
\label{app:prompts-en}

\subsection{Zero-shot (English)}
\label{app:prompts-zero-en}

\begin{PromptBox}
\begin{Prompt}
Task: Does the sentence below contain a “light verb construction (LVC)”?

Definition:
\begin{itemize}
    \item[\textbf{[1]}] If there is an LVC: the core predicational meaning is carried by a noun; the verb plays a “light/auxiliary” role, and the whole expression behaves as a single predicate.  
    Note: In this experiment, treat lexicalized/idiomatic noun+verb predicates as [1].
    \item[\textbf{[0]}] If there is no LVC: the verb is used with its basic (literal) meaning (e.g., a physical action, transfer of an object, literal seeing/hearing, etc.).
\end{itemize}

Sentence: ``{sentence}''

If the sentence contains an LVC, write only [1].  
If the sentence does not contain an LVC, write only [0].  

Do not provide any explanation; output only the label.

Answer:
\end{Prompt}
\end{PromptBox}

\subsection{One-shot}
\label{app:prompts-one}

\begin{PromptBox}{\textbf{One-shot}}
\begin{Prompt}
Task: Does the following sentence contain a “light verb construction (LVC)”?

Definition:
\begin{itemize}
    \item[\textbf{[1]}] If there is an LVC: A noun carries the core meaning of the predicate; the verb plays a more “light/auxiliary” role, and the construction functions as a single predicate.  
    Note: In this experiment, also count conventionalized/idiomatic noun+verb predicates as [1].
    \item[\textbf{[0]}] If there is no LVC: The verb is used with its basic literal meaning (physical action, object transfer, actual seeing/hearing, etc.).
\end{itemize}

Examples by verb:

\textbf{[VERB-\textit{DUY}]}

Sentence: ``He felt great respect for her/him.''  
Answer: [1]

Sentence: ``He heard the sound coming from the corridor.''  
Answer: [0]

\vspace{0.5em}
Now classify:

Sentence: ``{sentence}''

Write only [0] or [1]. Do not write anything else.

Answer:
\end{Prompt}
\end{PromptBox}

\subsection{Master few-shot (example; abbreviated)}
\label{app:prompts-master-example}

\begin{PromptBox}{\textbf{Master few-shot (example; abbreviated)}}
\begin{Prompt}
Task: Does the following sentence contain a “light verb construction (LVC)”?

Definition:
\begin{itemize}
    \item[\textbf{[1]}] If there is an LVC: A noun carries the core meaning of the predicate; the verb plays a more “light/auxiliary” role, and the construction functions as a single predicate.  
    Note: In this experiment, also count conventionalized/idiomatic noun+verb predicates as [1].
    \item[\textbf{[0]}] If there is no LVC: The verb is used with its basic literal meaning (physical action, object transfer, actual seeing/hearing, etc.).
\end{itemize}

Examples (abbreviated; representative example):

\textbf{[\textit{DUY-}]}

Sentence: ``For the first time in his/her life, he/she felt trust in someone.''  
Answer: [1]

Sentence: ``He/she heard the sounds coming from the corridor.''  
Answer: [0]

\medskip
\textbf{[\textit{VER-}]}

Sentence: ``He/she immediately gave an answer to my question.''  
Answer: [1]

Sentence: ``The teacher gave paper to the students.''  
Answer: [0]

\medskip
\textbf{[\textit{GIR-}]}

Sentence: ``When the exam started, he/she panicked.''  
Answer: [1]

Sentence: ``He/she quickly entered the room.''  
Answer: [0]

\medskip
... (other verbs and example pairs continue in the same format)

\vspace{0.5em}
Now classify:

Sentence: ``{sentence}''

Write only [0] or [1]. Do not write anything else.

Answer:
\end{Prompt}
\end{PromptBox}

\end{document}